\pdfoutput=1

\documentclass[11pt,a4paper]{article}
\usepackage[hyperref]{acl2021}
\usepackage{times}
\usepackage{latexsym}

\usepackage{defs}

\usepackage{microtype}

\aclfinalcopy 
\def\aclpaperid{2433} 

\title{The SelectGen Challenge: Finding the Best Training Samples for Few-Shot Neural Text Generation}

\author{ 
   Ernie Chang\thanks{ $\>$ Equal contribution. X.shen is now at Amazon Alexa AI.},  Xiaoyu Shen\footnotemark[1] ,  Alex Marin${}^{\Cap}$, Vera Demberg \\
   Dept. of Language Science and Technology, Saarland University
   \\  ${}^{\Cap}$ Microsoft Corporation, Redmond, WA \\

      {\tt \{cychang,xshen\}@coli.uni-saarland.de}
  \\ 
}

\date{}

\begin{document}
\maketitle
\begin{abstract}
We propose a shared task on training instance selection for few-shot neural text generation. Large-scale pretrained language models have led to dramatic improvements in few-shot text generation. Nonetheless, almost all previous work simply applies random sampling to select the few-shot training instances. Little to no attention has been paid to the selection strategies and how they would affect model performance. The study of the selection strategy can help us to (1) make the most use of our annotation budget in downstream tasks and (2) better benchmark few-shot text generative models. We welcome submissions that present their selection strategies and the effects on the generation quality.
\end{abstract}

\section{Introduction}
Few-shot text generation is an important research topic since obtaining large-scale training data for each individual downstream task is prohibitively expensive. Recently, pretraining large neural networks with a language
modeling objective has led to significant improvements across different few-shot text generation tasks~\cite{ radford2019language, lewis-etal-2020-bart} and many techniques are proposed based on them~\cite{chen2019few,schick2020few,zhang2020pegasus,kale2020text,chang2020dart,chang2021jointly,chang2021neural,li2021prefix}. However, all previous works simulate the few-shot scenario by randomly sampling a subset from the full training data. Little to no attention has been paid to the selection strategies. 
\begin{figure}[t]
  \centering
\includegraphics[width=\columnwidth]{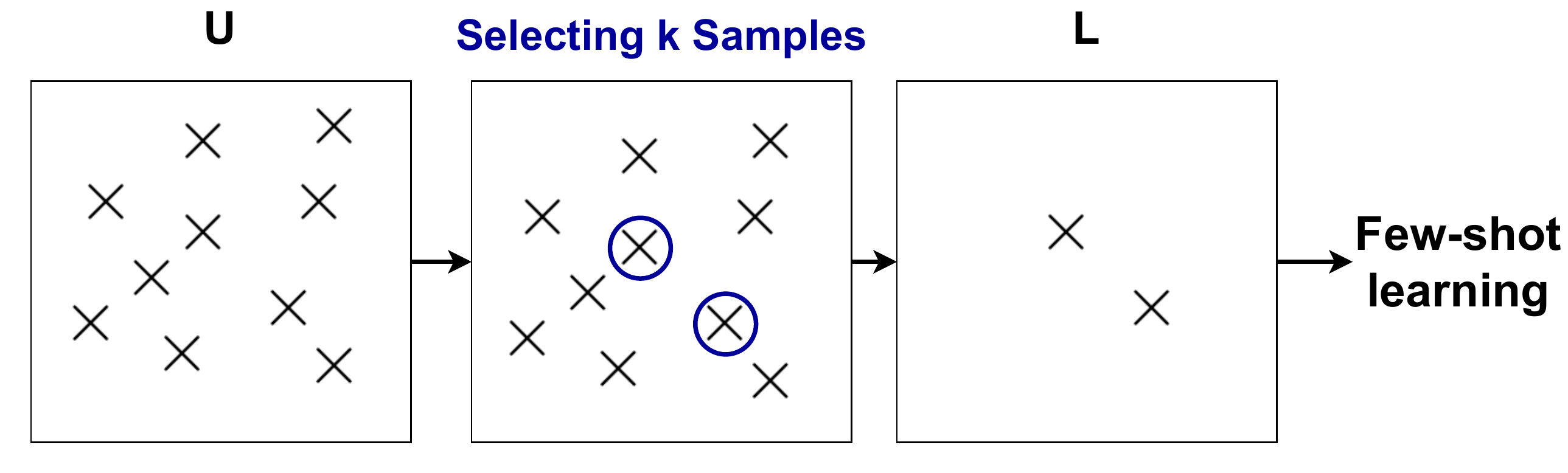}
\caption{ \small
\textbf{Training scenario for few-shot text generation}:  \textbf{U} represents unlabeled data and \textbf{L} indicates labeled instances. The annotation budget only allows selecting K data for annotating the reference text. We aim to identify the K most representative instances that, when annotated and trained on them, leads to a best model performance.
}
\label{fig:sample}
\end{figure}

The goal of the proposal is to call for innovative ideas on searching for an optimal strategy to select the few-shot training instances, as well as a comprehensive analysis of how the selection strategy would affect the model performance. 
The study of selection strategies is motivated by two rationales: 
First, random sampling leads to a large variance of model performance~\cite{zhang2020pegasus,schick2020few,schick2020s}. Yet current works sample their own training data which makes it difficult to compare across different models. One can then not be sure whether an improved performance can be really ascribed to the model or the randomness of sampling. Using a stable selection strategy to find the most informative few-shot instances can provide a fair platform and better benchmark different few-shot generative models. 
Second, in practical applications, e.g.~document summarization, the training data is usually obtained by manually annotating the summaries for some selected documents. In Figure~\ref{fig:sample}, we illustrate the typical training scenario for text generation where the annotation budget only allows annotating a limited amount of data. Studying the optimal selection strategy can help make the most use of our annotation budget. Specifically, we focus on the label-free setting where \emph{the selection can only condition on the unannotated data}. Although leveraging the reference text may benefit the selection strategy, it conflicts with the realistic setting where we need to first select the data then get its annotated reference text.

The selection task resembles the theme of active learning~\cite{balcan2007margin,chang2020dart,chang2021does,chang2021jointly}, where the model keeps identifying the most informative instances to get labeled. We can consider the task as a starting step before applying active learning, after which more annotations can be continuously collected to further improve the model.

\section{Task Description}
Following the training scenario shown in Figure~\ref{fig:sample}, we denote the unlabeled data as ${U_1, U_2,\ldots, U_n}$ where n is the data size. Depending on the downstream text generation task, ``data" can mean different types of inputs like unlabeled structured data, documents and paragraphs respectively in the context of data-to-text, document summarization and question generation. We will select $K$ instances from the whole unlabeled dataset, annotate them with reference text, and then train a neural generative model based on the annotated data. $K$ is defined based on the annotation budget. In this work, since we focus on the few-shot scenario, $K$ is set to be small ($\leq 2000$). The goal is to \emph{find the most representative $K$ instances that can lead to the optimal performance when being annotated and trained on them}.

\subsection{Submission Requirement}
Participants are required to submit: 
\begin{itemize}
    \item An executable code that takes as input a set of unlabeled data, outputs $K$ selected data that should be annotated.
    \item Selected training instances for $K=50, 200, 500$ and $2000$ together with model generations on the testset.
    \item A report that explains how the proposed selection strategy works and an analysis of its performance on the provided datasets.
\end{itemize}
 While it is acceptable to take into account task or language specific features, participants are encouraged to submit selection strategies that are:
\begin{itemize}
    \item Task agnostic. The selection strategy would work for a broad range of text generation tasks with various input-output formats.
    \item Language agnostic. The selection strategy can be seamlessly applied to same tasks in other languages.
    \item Model agnostic. The selection strategy can select most informative instances that improve the performance for a broad types of generative models (with various model architectures and training algorithms).
    \item $K$-agnostic. The selection strategy should work by varying the number of $K$.
\end{itemize}

\subsection{Data}
We will select representative datasets which cover three different text generation tasks. It will include but not limited to:
\begin{enumerate}
    \item Data-to-text: We use the dataset for the E2E challenge~\cite{novikova2017e2e} which contain 50,602 data-text pairs with 8 unique slots in the restaurant domain.
    \item Document Summarization: We use the CNN/Dailymail dataset (non-anonymized version)~\cite{hermann2015teaching} which contains 312,084 document-summary pairs.
    \item Question generation: We use the SQuAD dataset~\cite{rajpurkar2016squad} with over 100k questions. Following \citet{du2017learning}, we focus on the answer-independent scenario to directly generate questions from passages. 
\end{enumerate}

All the above datasets contain parallel input-output pairs for train/validation/test. We can simulate our few-shot scenario by only \emph{allowing leveraging $K$ input-output pairs from the training set}. The participants can decide which $K$ training instances to select based on all the inputs in the training set~\footnote{The submitted instance selection algorithm can only condition on the inputs in the training set. However, participants are welcome to incorporate the reference text or testset distribution to analyze the theoretical upper bound performance.}. Once the selected instances are determined, the model can then be trained on the $K$ input-output pairs. It is also worth mentioning that in order to simulate the true few-shot scenario, \emph{participants can only rely on the $K$ input-output pairs for both training and validation}, i.e., no extra held-out examples are available for hyperparameter tuning and model selection~\cite{schick2020few,perez2021true}. The participants can deside how to split them into the training and validation set.

We select the above three datasets only as examples for demonstration. Participants are encouraged to test their model on more diverse types of text generation tasks, e.g., tasks from the GEM benchmark~\cite{gehrmann2021gem}. Nevertheless, we recommend participants to first test and analyze their model on the above three datasets. In the final test, we will evaluate on the above three datasets to allow comparison across different submission. It is, however, totally acceptable to not target at all of the above three tasks. The participants can decide the tasks and datasets depending on their interest.

\subsection{Generative Model}
It is encouraged that participants can test their selection strategy on a wide list of generative models. In the end, to allow for a fair comparison across all submissions, we will test the selection algorithm by finetuning the open-sourced Bart model~\cite{lewis-etal-2020-bart} on the selected training instances with maximum likelihood. Bart is pretrained with a denoising autoencoder objective on large amount of text data and has been the state-of-the-arts for many text generation tasks. Therefore, we recommend to first test with this standard generative model. There have been many algorithms proposed for improved generation quality under the few-shot scenario like pattern exploitation training~\cite{schick2020few,li2021prefix,lester2021power} and cyclic training~\cite{tseng2020generative,chang2021jointly,guo2021fork}. We welcome test results using different types of generative models. Nonetheless, \emph{the focus of the shared task is on the instance selection algorithm but not the few-shot generative model}. While it is nice to provide data points that demonstrate state-of-the-art results, generating with the most advanced model for better evaluation scores is by not means the main purpose.

\subsection{Schedule}
We follow the following schedule for the shared task of training instance selection:
\begin{itemize}
    \item \textbf{December 15th, 2021}. The shared task is
announced along with the selected text generation tasks and datasets.
    \item \textbf{February 15th, 2022}. The submission system and public leaderboard are open. Participants can deploy and test models with the provided automatic evaluation scripts.
    \item \textbf{May 15th, 2022}. This is the deadline for software and report submission. The manual evaluation begins. We will test the submitted selection algorithms with the same generative model and hyperparameter tuning mechanism. Model outputs will be compared with both automatic metrics and human evaluation.
    \item \textbf{June 15th, 2022}. The results of the automatic metrics and human evaluations will be announced.
\end{itemize}

After getting all the evaluation results, we will make a report to analyze different submissions. The shared task’s findings are
then presented at the following INLG.
\section{Evaluation}
The final evaluation will be conducted on the following two settings:
\begin{enumerate}
    \item We apply the submitted selection algorithms to select $K$ training instances and then finetune on them using a fixed strategy (with Bart model, same train/validation split and hyperparameter tuning mechanisms). The purpose is to evaluate all selection algorithms under a fair setting. In this setting, we will run the selection algorithm and training pipeline on our side to ensure fairness.
    \item For each submission, we evaluate the model outputs of the best system. The purpose is to get an upper bound score for few-shot text generation with the best combination of settings (random seed, generative model, optimization algorithm, train/validation split, hyperparameter tuning, etc). In this setting, we will rely on the submissions of model outputs from the participants.
\end{enumerate}

We will provide scripts for the automatic evaluation. The human evaluation will be conducted after all submissions are received under the same platform and metrics.
\subsection{Automatic Evaluation}
The evaluation metrics differ according to the downstream tasks. The metrics used for the final evaluation will be announced after the submission system is open. Participants are encouraged not to focus on one specific metric to avoid overfitting to it. The final evaluation will adopt metrics following into the following categories:
\begin{itemize}
    \item Lexical similarity, which measure the lexical overlap between the model output and the gold references, including many popular metrics like BLEU~\cite{papineni2002bleu}, ROUGE~\cite{lin2004rouge} and METEOR~\cite{banerjee2005meteor}.
    \item Semantic relevance, which measures the semantic similarity between the model output and the gold references, including the newly proposed BertScore~\cite{zhang2019bertscore} and BLEURT~\cite{sellam2020bleurt}.
    \item Consistency with task input, which measures if the output contains consistent information with the task input and no hallucinations. Many works have proposed metrics based on question answering~\cite{eyal2019question,durmus2020feqa}, natural language inference~\cite{kumar2020nile} and mutual information~\cite{shen2018nexus,zhang2018generating}.
    \item Output diversity, which measures if the model can produce diverse outputs with different inputs, including metrics like the count and entropy of distinct uni/bi-grams~\cite{li2016diversity,duvsek2020evaluating}. 
    \item Other task-specific requirement, e.g., slot-error rate for data-to-text and compression rate for document summarization.
\end{itemize}
After the submission system opens, we will announce the metrics we picked for the automatic evaluation and provide the evaluation script.
\subsection{Human Evaluation}
We will also provide human evaluation scores on the system outputs since none of the automatic metrics can correlate perfectly with the generation quality. We will follow the recently proposed taxonomy of human evaluation measures by ~\citet{belz-etal-2020-disentangling,su2020moviechats} and follow
the reporting strategies proposed by ~\citet{howcroft2020twenty}. The human evaluation will be focused on the following two parts, which are specifically hard to be accurately measured by automatic metrics:
\begin{itemize}
    \item Fluency. If the output itself is a fluent sentence that can be well understood by humans, defined by a 5-scale Likert score.
    \item Consistency. If the output is consistent with the input and does not contain hallucinations, defined by a binary true/false score.
\end{itemize}

The human evaluation will be conducted after collecting all the submissions. It will be performed under a unified pipeline and annotation guideline to make sure results are comparable across model outputs from all submitted systems. To make the analysis comprehensive, participants are nonetheless encouraged to also perform their own human evaluation and include the results in their report.
\subsection{Variance of Model}
An important factor worth mentioning is the variance of the model. The variance of the model output can come from different steps, e.g., variance of the selection algorithm, random seed of training, hyperparameter selection, etc. It is rather straightforward to simply apply a random sampling strategy to select the $K$ training instances and find a relatively good selection choice by brute force. However, this is clearly against the purpose of the shared task. We aim to find out a \emph{selection algorithm that can stably help us identify the most representative training instances} instead of only getting the instance set. Therefore, when doing the final evaluation, if the submitted selection algorithm is not deterministic, we will run the algorithm 5 times to get 5 different selection sets and aggregate the results. The variance of the evaluation will also be reported (For the setting 1 of evaluation). For setting 2, we rely on the participants themselves to provide the selected instance set and the model outputs. Participants must indicate clearly how the instance set is determined, e.g., whether they cherry-pick a best instance set by randomly running the algorithm for many times, or leverage other information like the reference text for other inputs, testset distribution, etc.
\section{Conclusion}
In this proposal, we target at the problem of training instance selection for few-shot text generation. Current research simply applies random sampling which has a large selection variance and can lead to suboptimal performance. The main goal of the task is to call for more attention on this largely under-explored problem, gather innovative ideas on proposing selection algorithms and provide a fair platform for comparison. 

We believe our shared task can be an important supplement to the study of few-shot text generation, where most works focus solely on the generative algorithm while neglecting the training instance selection. Selection strategies proposed in this task can be used to better benchmark model performances for few-shot text generation.
Importantly, the task was inspired by realistic industrial settings and requirements and will hopefully benefit multiple areas of NLP research including human-in-the-loop learning and other active learning based research, where the resource and time constraints calls for the task to be performed.

\section*{Acknowledgements}
This research was funded in part by the German Research Foundation (DFG) as part of SFB 248 ``Foundations of Perspicuous Software Systems''. We sincerely thank the anonymous reviewers for their insightful comments that helped us to improve this paper. 

\bibliographystyle{acl_natbib}
\bibliography{anthology,acl2021}

\end{document}